\documentclass[sigconf]{acmart}
\AtBeginDocument{%
  }
\setcopyright{acmlicensed}
\copyrightyear{2026}
\acmYear{2026}
\setcopyright{cc}
\setcctype{by}

\acmConference[WSDM '26] {Proceedings of the Nineteenth ACM International Conference on Web Search and Data Mining}{February 22--26, 2026}{Boise, ID, USA.}
\acmBooktitle{Proceedings of the Nineteenth ACM International Conference on Web Search and Data Mining (WSDM '26), February 22--26, 2026, Boise, ID, USA}
\acmISBN{979-8-4007-2292-9/2026/02}
\acmDOI{10.1145/3773966.3777931}
\usepackage{enumitem}
\usepackage{multirow}
\usepackage{algorithm}
\usepackage{algorithmic}
\usepackage{newfloat}
\usepackage{listings}
\usepackage{booktabs}
\usepackage{amsfonts}
\usepackage{colortbl}
\usepackage{amsmath}
\usepackage{xcolor}

\begin{document}
\title{Can Slow-Thinking LLMs Reason Over Time? Empirical Studies in Time Series Forecasting}

\author{Mingyue Cheng}
\affiliation{%
  \institution{State Key Laboratory of Cognitive Intelligence,  University of Science and Technology of China}
  \city{Hefei}
  \state{Anhui Province}
  \country{China}
}
\email{mycheng@ustc.edu.cn}

\author{Jiahao Wang}
\affiliation{%
  \institution{State Key Laboratory of Cognitive Intelligence,  University of Science and Technology of China}
  \city{Hefei}
  \state{Anhui Province}
  \country{China}
}
\email{jiahao.wang@mail.ustc.edu.cn}

\author{Daoyu Wang}
\affiliation{%
  \institution{State Key Laboratory of Cognitive Intelligence,  University of Science and Technology of China}
  \city{Hefei}
  \state{Anhui Province}
  \country{China}
}
\email{wdy030428@mail.ustc.edu.cn}

\author{Xiaoyu Tao}
\affiliation{%
  \institution{State Key Laboratory of Cognitive Intelligence,  University of Science and Technology of China}
  \city{Hefei}
  \state{Anhui Province}
  \country{China}
}
\email{txytiny@mail.ustc.edu.cn}

\author{Qi Liu}
\authornote{Qi Liu is the corresponding author.}
\affiliation{%
  \institution{State Key Laboratory of Cognitive Intelligence,  University of Science and Technology of China}
  \city{Hefei}
  \state{Anhui Province}
  \country{China}
}
\email{qiliuql@ustc.edu.cn}

\author{Enhong Chen}
\affiliation{%
  \institution{State Key Laboratory of Cognitive Intelligence,  University of Science and Technology of China}
  \city{Hefei}
  \state{Anhui Province}
  \country{China}
}
\email{cheneh@ustc.edu.cn}

\renewcommand{\shortauthors}{Mingyue Cheng et al.}

\begin{abstract}
Time series forecasting (TSF) traditionally relies on fast-thinking paradigms that map historical observations directly to future sequences of continuous values. While effective, such approaches often frame forecasting as a pattern-matching problem and tend to overlook explicit reasoning over temporal dynamics and contextual factors, which are critical for modeling long-range dependencies and non-stationary behaviors in real-world scenarios. Recent slow-thinking large language models (LLMs), such as OpenAI o1 and DeepSeek-R1, demonstrate strong inference-time multi-step reasoning abilities. This raises a fundamental question: can slow-thinking LLMs reason over temporal dynamics to support accurate TSF, even without task-specific training? To investigate this question, we present TimeReasoner, a systematic empirical study that reformulates TSF as a conditional reasoning process performed entirely at inference time. TimeReasoner integrates hybrid instructions consisting of task directives, timestamps, sequential values, and optional contextual features, and induces multi-step temporal reasoning in pretrained slow-thinking LLMs through chain-of-thought prompting and rollout-based reasoning strategies. Extensive experiments across diverse TSF benchmarks show that slow-thinking LLMs consistently outperform prior baselines or achieve competitive training-free forecasting performance. Beyond accuracy, we analyze how different inference-time reasoning strategies influence forecasting behaviors, highlighting both the potential and limitations of slow-thinking paradigms for TSF.\footnote{\url{https://github.com/realwangjiahao/TimeReasoner}}

\end{abstract}

\ccsdesc[500]{Mathematics of computing~Time series analysis}

\keywords{Forecasting, Long Chain-of-thought, Time Series Reasoning}
\maketitle


\vspace{-0.1in}

\section{Introduction}
Time series forecasting (TSF) is a fundamental data mining task with wide-ranging applications in domains such as finance~\cite{tay2001application}, energy~\cite{deb2017review}, and healthcare~\cite{kaushik2020ai} where accurate forecasts are critical for strategic planning and effective decision-making under uncertainty~\cite{box1968some,pan2025onecast}. The goal of TSF is to predict future values of target variables by leveraging historical observations together with contextual features, including temporal covariates and static attributes~\cite{hu2016building,wang2024tabletime,jintaozhang}.

Over the past decades, numerous approaches have been proposed for time series forecasting (TSF)~\cite{cheng2025comprehensive,NEURIPS2023_2e19dab9}, which can be broadly categorized into traditional statistical modeling approaches~\cite{lim2021time} and data-driven approaches. Traditional statistical models, such as ARIMA~\cite{zhang2003time} and Holt–Winters~\cite{chatfield1978holt}, rely on strong assumptions of linearity, stationarity, and seasonality. While offering clear explainability through explicit model structures, they often struggle to capture complex, nonlinear, or non-stationary temporal patterns. In contrast, data-driven approaches, particularly deep learning–based models, capture intricate temporal patterns such as trends and seasonality through nonlinear architectures, achieving strong performance across standard benchmarks~\cite{benidis2022deep,zhang2024trajectory} and demonstrating further improvements in various TSF scenarios~\cite{NEURIPS2024_053ee34c,ilbert2024analysing}. With recent advances in large language models (LLMs), TSF has also been explored from multimodal understanding and generative modeling perspectives, leveraging the ability of LLMs to incorporate heterogeneous contextual features~\cite{jiang2024empowering}. Subsequent studies extend this line of work through large-scale LLM integration and generation-oriented architectures~\cite{zhang2024large}. Despite these advances, most existing LLM-based TSF approaches still treat forecasting as a direct sequence mapping problem, producing predictions in a single inference pass without explicit intermediate reasoning. Consequently, the potential of LLMs for step-wise temporal reasoning—an ability that may be critical for long-horizon and context-aware forecasting—remains largely under-explored~\cite{tan2024language,gat2023faithful}.

Building on this observation, we further abstract existing TSF methods under a unified perspective and note that most of them largely follow a fast-thinking paradigm~\cite{jha2016comparative}. Specifically, forecasting is typically performed through primarily single-step inference, mapping historical inputs directly to future targets with limited explicit intermediate reasoning~\cite{chow2024towards}. Under this paradigm, models often capture short-term temporal dependencies effectively, while systematically integrating heterogeneous contextual information in a step-by-step manner remains challenging~\cite{kauppinen2007modeling}. In contrast, slow-thinking LLMs, such as OpenAI o1~\cite{o1} and DeepSeek-R1~\cite{guo2025deepseek}, demonstrate strong multi-step reasoning abilities in domains such as mathematics, typically by generating explicit intermediate reasoning steps before producing final answers~\cite{muennighoff2025s1}. Such reasoning processes may offer advantages for modeling long-range dependencies and contextual regime shifts in TSF. However, it remains under-explored whether these reasoning capabilities can be effectively leveraged for time series forecasting, particularly in training-free settings without task-specific fine-tuning.

These questions, centered on the potential of slow-thinking LLMs for TSF, motivate a re-examination of how forecasting problems are formulated and solved. In this work, we investigate whether slow-thinking LLMs can function as forecasters through explicit, inference-time step-by-step reasoning. Specifically, we reformulate TSF as a training-free conditional reasoning task, in which predictions are guided by historical observations together with various contextual features. Accordingly, we propose TimeReasoner, a inference-time reasoning framework that leverages LLMs’ semantic reasoning capabilities to produce both interpretable forecasts and structured reasoning traces. We further analyze how different reasoning strategies, prompt construction, and modeling choices influence forecasting performance, and investigate the following research questions:

\noindent \textbf{RQ1: How does TimeReasoner perform compared to state-of-the-art baselines?}
We evaluate the forecasting performance of TimeReasoner across multiple benchmarks to establish its overall effectiveness relative to existing methods.  

\noindent \textbf{RQ2: Which key settings in hybrid instruction and reasoning strategies are essential for accurate forecasting?}  
Through ablation on TimeReasoner’s key configurations—including hybrid prompt components (timestamps, contextual features, raw series) and alternative reasoning strategies—we identify the factors that most influence predictive accuracy.  

\noindent \textbf{RQ3: How trustworthy is TimeReasoner under real-world input imperfections?} We simulate incomplete and noisy inputs by introducing missing values and perturbations, evaluating the model’s ability to maintain accuracy and reasoning quality under such conditions. We further quantify its capability to estimate and communicate forecasting uncertainty.  

\noindent \textbf{RQ4: How does TimeReasoner perform under different TSF task settings?}  
We evaluate the model’s forecasting accuracy and reasoning trace across varying look-back windows and prediction lengths to understand its adaptability to different temporal settings.

\noindent \textbf{RQ5: How do inference-time hyperparameters affect forecasting quality?}  
We analyze the sensitivity of TimeReasoner to key hyperparameters such as temperature and base LLM selection, examining their impact on prediction accuracy.  

\noindent \textbf{RQ6: Do reasoning traces from TimeReasoner provide explainable insights into its predictions?}  
We examine intermediate reasoning steps and representative case studies to assess their  consistency with observed input patterns, and potential to enhance the explaination of the model’s forecasts.  

\begin{figure*}[th]
	\centering
	\includegraphics[width=1\linewidth]{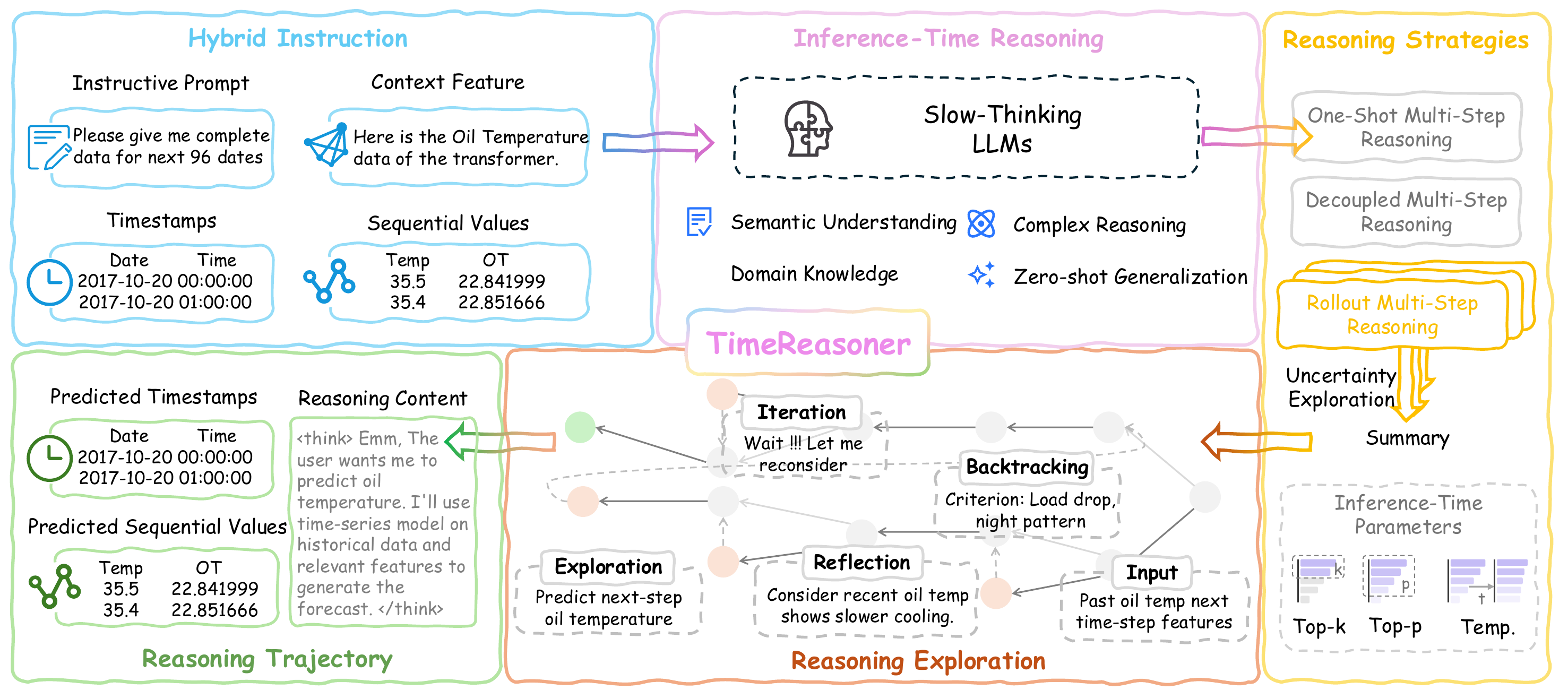}
	\caption{Overall framework of TimeReasoner for training-free reasoning-based time series forecasting.}
    \Description{Overall framework of TimeReasoner for training-free reasoning-based time series forecasting.}
	\label{fig:framework}
\end{figure*}

\vspace{-0.12in}
\section{Related Work}
In this section, we briefly introduce the related work, including time series forecasting and LLM-based reasoning.
\vspace{-0.1in}
\subsection{Time Series Forecasting}
Time series forecasting has been extensively studied over the past decades~\cite{cheng2025comprehensive}.  
Early statistical methods, such as ARIMA~\cite{box1968some,zhang2003time}, relied on auto-regressive and moving-average formulations with strong theoretical guarantees, but often assumed ideal data properties that rarely hold in practice~\cite{hewamalage2021recurrent}.  
With the growth of data and computing power, deep learning-based sequence-to-sequence models have emerged~\cite{salinas2020deepar}.  
Recurrent neural networks were among the first to capture temporal dependencies~\cite{wen2017multi}, but suffered from limited receptive fields and error accumulation in recursive inference~\cite{lai2018modeling}.  
Subsequent advances introduced architectures for long-range dependency modeling~\cite{zhou2021informer}, including self-attention and convolutional designs~\cite{li2019enhancing,liu2022scinet}.  
Recent work has also integrated classical techniques—such as trend-seasonal decomposition and time-frequency transforms~\cite{zhou2022fedformer,liu2024generative}—into neural networks~\cite{wen2020fast,zhou2022film,zhang2025alphacast}.  
Notably, even simple linear models enhanced with decomposition can be competitive~\cite{zeng2023transformers}, and slice-based approaches achieve strong performance in long-term forecasting~\cite{nie2022time,zhang2023crossformer,cheng2025instructime}.

In recent years, large language models (LLMs) have attracted considerable attention for their ability to understand and generate human-like text~\cite{kumar2024large,yao2024survey,tao2024hierarchical}.  Beyond their success in natural language processing~\cite{zhang2024large,tao2025values}, LLMs have shown strong potential in time series analysis~\cite{jin2024position}.  Their application to time series generally follows two paradigms: \emph{fine-tuning} and \emph{prompt-based zero-shot learning}~\cite{jiang2024empowering}.   In fine-tuning, a pre-trained LLM is further adapted to time series tasks by training on task-specific data~\cite{chang2023llm4ts,chang2024align}, enabling it to capture domain patterns and improve performance~\cite{liu2025calf,jin2023time}, albeit with substantial labeled data and computational costs~\cite{cheng2025comprehensive}.   Prompt-based zero-shot methods instead guide predictions via well-designed prompts without task-specific training~\cite{gruver2023large,liu2024lstprompt}, offering greater flexibility and efficiency~\cite{tang2025time} but sometimes underperforming fine-tuned models in specialized scenarios~\cite{wang2024tabletime}.   Both paradigms underscore the rising interest in LLMs for time series, while revealing challenges in optimizing their effectiveness for such tasks.

\subsection{LLM-based Reasoning}
LLM-based reasoning refers to leveraging large language models’ ability to perform multi-step, structured thinking for complex tasks.  
Within this paradigm, \textit{inference-time reasoning}~\cite{zeng2025revisiting} has emerged as an effective strategy to enhance a pre-trained model’s performance on challenging reasoning problems by allocating additional computational resources during inference~\cite{singhi2025solve,yang2025towards}.  
These resources may include extended processing time, increased computational steps, or advanced strategies such as multi-path sampling and iterative refinement~\cite{leehanchung}.  
In contrast to training-time scaling—which improves model capability through more parameters, data, or computation during training~\cite{parashar2025inference}—inference-time reasoning exploits extra computation at inference to activate deeper reasoning potential~\cite{zhang2025and} without modifying model weights~\cite{chen2024simple}.  
Recent LLMs demonstrate the effectiveness of this approach: DeepSeek-R1~\cite{guo2025deepseek} applies iterative refinement to boost coding and reasoning performance~\cite{mercer2025brief}; Gemini-2.5-pro~\cite{team2023gemini} is expected to excel on sophisticated tasks with extended processing or multi-step protocols; and OpenAI’s GPT-o1~\cite{hasei2025comparative} achieves significant gains on challenging reasoning benchmarks using chain-of-thought prompting~\cite{wei2022chain}.  
These examples underscore the growing role of inference-time reasoning in trading inference-time computation for higher accuracy~\cite{muennighoff2025s1}.

Despite the transformative progress slow-thinking reasoning has brought to numerous complex reasoning tasks, whether such powerful reasoning models can be effectively leveraged to improve time series forecasting has yet to be explored.

\vspace{-0.2in}
\section{Problem Definition}
Given a multivariate time series \( \mathcal{Y} = \{ \mathbf{y}_t \}_{t=1}^T \), where \( \mathbf{y}_t \in \mathbb{R}^d \) denotes the \( d \)-dimensional target variables at time step \( t \), the goal of multivariate time series forecasting (TSF) is to predict future values \( \{\mathbf{y}_{t+1}, \ldots, \mathbf{y}_{t+H} \} \) over a fixed predict window \( H \). At each forecasting step \( t \), the model is provided with a lookback window of length \( L \), i.e., \( \{\mathbf{y}_{t-L}, \ldots, \mathbf{y}_{t-1}\} \), summarizing recent historical observations. In addition, the model may access a context vector \( \mathcal{Z} \in \mathbb{R}^k \) containing supplementary information such as temporal covariates, static attributes, or other relevant features to aid forecasting.


\vspace{-0.1in}
\section{The Proposed TimeReasoner}
In this section, we present the overall architecture of TimeReasoner, detailing how it leverages LLMs to perform  inference-time reasoning for time series forecasting.

\vspace{-0.1in}
\subsection{Overview of TimeReasoner}
Figure~\ref{fig:framework} illustrates the overall architecture of TimeReasoner, a training-free, inference-time reasoning framework for time series forecasting using slow-thinking LLMs. The framework starts with Hybrid Instruction, where instructive prompts are integrated with contextual features (e.g., timestamps, sequential values) to construct structured inputs. These inputs are processed in the Inference-Time Reasoning stage, where slow-thinking LLMs leverage semantic understanding, domain knowledge, and multi-step reasoning capabilities to generate forecasts without task-specific fine-tuning.
To further improve forecasting accuracy, reasoning strategies including one-shot multi-step, decoupled multi-step, and rollout multi-step reasoning are employed, along with uncertainty exploration and inference-time parameter control. Finally, Reasoning Exploration records the reasoning trajectories and supports iterative refinement, backtracking, reflection, and explanation, thereby producing accurate predictions accompanied by interpretable reasoning traces.

\vspace{-0.1in}

\subsection{Hybrid Instruction}
To support time series reasoning, we design multimodal prompts including hybrid instructions that combine:
\begin{itemize}[leftmargin=*]
\item \textit{Instructive prompts.}
TimeReasoner utilizes an instructive prompt to explicitly specify the forecasting objective, such as the number of future time steps to be predicted. This directive helps the LLM focus on the target prediction range and aligns its generative reasoning with the user’s intent.
\item \textit{Contextual features.}  
TimeReasoner is provided with natural language descriptions that offer contextual information like domain-level knowledge and channel-specific semantics. These help align numerical data with real-world meaning, enhancing explainability and task relevance~\cite{williams2024context}.
\item \textit{Raw time series in original space.}  
TimeReasoner takes time series in their original scale as input, without normalization or standardization. Preserving the raw magnitudes allows the model to reason over absolute values and real-world fluctuations that might otherwise be distorted.
\item \textit{Timestamps.}  
TimeReasoner incorporates original timestamps into the prompt, enabling it to recognize both absolute temporal positions (such as specific hours or dates) and relative patterns (such as periodicity), which are crucial for temporal reasoning~\cite{wang2024rethinking,zeng2024much}.
\end{itemize}

\subsection{Inference-Time Reasoning}
Based on the constructed multimodal prompts, TimeReasoner performs inference-time reasoning by leveraging the semantic understanding and complex reasoning capabilities of slow-thinking LLMs. Specifically, the model first interprets the forecasting objective and contextual descriptions to establish a high-level semantic grounding of the task, including the prediction horizon, domain constraints, and plausible temporal behaviors. It then reasons over the raw time series values and timestamps jointly, identifying salient temporal patterns such as trends, periodicity, regime shifts, and potential anomalies directly in the original value space. During generation, the LLM incrementally integrates these observations, reflecting on recent dynamics and historical regularities to form a coherent forecasting trajectory. This inference-time reasoning process allows TimeReasoner to adaptively combine numerical evidence with domain semantics, enabling flexible and interpretable time series forecasting without task-specific parameter tuning.

\vspace{-0.1in}
\subsection{Reasoning Strategies}
We explore three distinct reasoning paradigms to evaluate the forecasting capability of slow-thinking LLMs.
\begin{itemize}[leftmargin=*]
\item \textit{One-Shot Reasoning.}
TimeReasoner performs a single, comprehensive reasoning pass on the given input. During the generation, it processes all necessary information and internally executes the multi-step logic required. The model then outputs the complete set of predicted results, without any subsequent iteration.
\item \textit{Decoupled Reasoning.}
Unlike one-shot reasoning, the reasoning process is decoupled: the model generates a partial thought, reflects on it, and then continues, forming a generate–reflect–generate loop within a single inference, enabling multi-step reasoning.
\item \textit{Rollout Reasoning.}
In rollout reasoning, TimeReasoner generates predictions incrementally, producing one part of the result in each step. This process involves several iterative rounds, where each output serves as context for the next, enabling TimeReasoner to gradually build toward a comprehensive final result.
\end{itemize}

\vspace{-0.1in}
\subsection{Reasoning Exploration and Trajectory}
Beyond directly generating numerical forecasts, TimeReasoner leverages the slow-thinking capability of LLMs to perform iterative reasoning over time series dynamics at inference time. 
Given historical observations and timestamps, the model incrementally analyzes recent temporal behaviors, such as local trends, periodic patterns, and potential regime changes, and reflects on how these dynamics may evolve over the predict window. 
During this process, the LLM may revisit earlier hypotheses, backtrack from inconsistent assumptions, and refine its interpretation of the time series when new evidence (e.g., abrupt fluctuations or weakened seasonality) is identified. 
This iterative reasoning and reflection enable the model to form a coherent forecasting rationale rather than relying on a single-pass mapping from past to future values.

To obtain a stable and reliable prediction, we execute the slow-thinking forecasting procedure multiple times under the same input and aggregate the generated results; specifically, we perform three independent generations and take the mean of the resulting forecast sequences as the final output. 
This aggregation strategy mitigates variability introduced by stochastic inference and provides a more robust estimate of future time series values, while preserving the explainability afforded by explicit reasoning traces.

\vspace{-0.1in}
\section{Experimental Settings}
In this section, we describe the experimental settings for evaluating TimeReasoner, covering datasets, baselines, implementation details.

\vspace{-0.1in}
\subsection{Datasets}
\begin{table}[t!]
	\centering
	\caption{Statistical information of experimental datasets.}
    \vspace{-0.12in}
	\resizebox{\columnwidth}{!}{  
		\begin{tabular}{ccccc}
			\hline
			Dataset & Domain & Length & Features & Frequency \\ \hline
			ETTh1\&ETTh2 & Electricity & 17,420 & 7 & 1 hour \\ 
			ETTm1\&ETTm2 & Electricity & 69,680 & 7 & 15 mins \\
			AQWan\&AQShunyi & Environment & 35,064 & 11 & 1 hour  \\ 
			Exchange & Economy & 7,588 & 8 & 1 day  \\ 
			NASDAQ & Economy & 1,244 & 5 & 1 day \\ 
			Wind & Energy & 48,673 & 7 & 15 mins \\ 
			VitalDB & Healthcare & 2,855 & 2 & 3s \\ \hline
		\end{tabular}
	}
    \vspace{-0.2in}
	\label{dataset}
\end{table}

To evaluate the generalization of our approach across diverse scenarios, we deliberately select datasets from multiple domains rather than a few popular benchmarks. The datasets include: ETT (electricity load, 2016–2018)~\cite{zhou2021informer}; Exchange (foreign exchange rates, 1990–2016)~\cite{lai2018modeling}; Wind (wind measurements, 2020–2021)~\cite{lai2018modeling}; AQ (air quality, four years)~\cite{zhang2017cautionary}; NASDAQ (stock market prices, volumes, and high–low values)~\cite{feng2019temporal}; and VitalDB (physiological signals sampled every 3 seconds)~\cite{cheng2024hmf}. These datasets span energy, finance, climate, environmental monitoring, and healthcare, offering heterogeneous temporal patterns, sampling rates, and contextual features. Such diversity supports robust reasoning-based forecasting evaluation beyond narrow benchmark settings and reduces bias toward any single data distribution. Following standard practice, each dataset is chronologically split into training, validation, and testing sets. Table~\ref{dataset} provides detailed descriptions.
\begin{table*}[t]
\centering
\caption{Performance comparison of TimeReasoner and baseline models with best values in \textbf{bold} and second-best \underline{underlined}. MSE $\downarrow$ and MAE $\downarrow$ are used as the evaluation metrics. Overall, TimeReasoner achieves superior performance across datasets.}
\vspace{-0.12in}
\label{main_result}
\renewcommand{\arraystretch}{1.3}
\resizebox{\textwidth}{!}
{
\large
\begin{tabular}{lc|cccccccccc}
\toprule
\textbf{Methods}&\textbf{Metric}&\textbf{ETTh1}&\textbf{ETTh2}&\textbf{ETTm1}&\textbf{ETTm2}&\textbf{Exchange}&\textbf{AQWan}&\textbf{AQShunyi}&\textbf{Wind}&\textbf{NASDAQ}&\textbf{VitalDB}\\
\hline
\multirow{2}{*}{DLinear}&MSE&7.7656 & 10.3584 & \underline{14.0352} & 7.8604 & 0.0014 & 21090.0523 & 21077.8247 & 1617.3343 & \textbf{0.0007} & 68.2254\\
&MAE&1.4981 & 2.0567 & 1.7591 & 1.8981 & 0.0253 & 51.1701 & 50.5644 & 18.3691 & 0.0215 & 6.7148\\
\hline
\multirow{2}{*}{PatchTST}&MSE&9.2430 & 10.9489 & 16.3988 & 5.6375 & \underline{0.0009} & \underline{12528.5571} & 16747.3794 & 2013.9736 & \textbf{0.0007} & \underline{50.7402}\\
&MAE&1.6500  & 1.9930 & 1.9294 & \textbf{1.3834} & 0.0198 & 39.8232 & 42.0165 & 20.1837 & \textbf{0.0211} & \textbf{5.6983}\\
\hline
\multirow{2}{*}
{iTransformer}&MSE&7.5195 & 9.9716 & \textbf{12.7432} & 5.7392 & 0.0010 & 13577.9277 & 18147.8075 & \underline{1600.0376} & 0.0008 & 79.2864 \\
&MAE&1.4983 & 1.9010 & \textbf{1.6374} & 1.4422 & 0.0202 & 39.8069 & 42.7643 & \underline{17.9244} & 0.0235 & 7.3322 \\
\hline
\multirow{2}{*}
{TimeXer}&MSE&8.4878 & 11.4123 & 14.0473 & \textbf{5.5801} & \underline{0.0009} & 14509.0913 & 16505.3908 & 1668.6824 & \textbf{0.0007} & 65.8567 \\
&MAE&1.5430 & 2.0606 & 1.7534 & 1.4238 & 0.0193 & 41.2693 & \underline{40.9784} & 18.2386 & 0.023 & 6.1253 \\
\hline
\multirow{2}{*}
{GPT4TS}&MSE&6.9454 & \underline{9.7156} & 15.9028 & \underline{5.6327} & \underline{0.0009} & 13543.6153 & 16821.9321 & 1786.3085 & 0.0010 & 65.3715 \\
&MAE&1.4127&1.8891&1.9364&1.4743&0.0200&39.7527&41.8708&18.2427&0.0253&6.7207\\
\hline
\multirow{2}{*}
{Time-LLM}&MSE&\underline{6.8152}&9.9876&15.8078&5.6787&0.0010&13405.9845&16678.0982&1773.9651&0.0011&65.9761\\
&MAE&\underline{1.4115}&\underline{1.8880}&1.9350&1.4733&0.0199&\underline{39.7481}&41.8598&18.2365&0.0252&6.7151\\
\hline
\multirow{2}{*}
{LLMTime}&MSE&10.6261&11.2011&14.6394&6.9001&0.0026&29141.5143&30266.3043&3979.7996&0.0021&99.5761\\
&MAE&1.6987&1.5876&1.9543&1.5109&0.0345&65.6789&60.3456&30.1234&0.0234&8.9054\\
\hline
\multirow{2}{*}
{Chronos}&MSE&10.6328 & 15.2152 & 22.7706 & 9.6567 & 0.0103 & 13858.3941 & 16601.0443 & 2874.5295 & 0.0043 & \textbf{50.2799}\\
&MAE&1.5128 & 2.1104 & 2.2046 & 1.7979 & 0.0324 & 42.7446 & 45.1168 & 23.1097 & 0.0341 & \underline{5.7185}\\
\hline
\multirow{2}{*}
{Moirai}&MSE&10.5981 & 11.9302 & 38.5828 & 10.4213 & \textbf{0.0008} & 14864.1762 & 19450.7758 & 2279.2915 & 0.0009 & 180.8878\\
&MAE&1.5690 & 1.9536 & 2.7939 & 1.7779 & \underline{0.0182} & 42.0587 & 44.1069 & 20.0049 & 0.0245 & 10.6444 \\
\hline
\multirow{2}{*}
{MOMENT}&MSE&18.3100 & 12.1693 & 26.2155 & 8.1986 & 0.0012 & 14687.9019 & \underline{15596.0312} & 1726.4521 & 0.0006 & 73.3861\\
&MAE&2.3847 & 2.1208 & 2.8222 & 1.7389 & 0.0223 & 43.5303 & 43.4243 & 20.0301 & 0.0201 & 6.8837\\
\hline
\multirow{2}{*}
{TimeReasoner}&MSE&\textbf{5.4469}&\textbf{8.6020}&14.2055&6.4384&\underline{0.0009}&\textbf{11305.6345}&\textbf{12874.9412}&\textbf{1556.5316}&\underline{0.0008}&79.4890\\
&MAE&\textbf{1.1984}&\textbf{1.6811}&\underline{1.6984}&\underline{1.4209}&\textbf{0.0168}&\textbf{36.1614}&\textbf{37.0193}&\textbf{17.5034}&\underline{0.0215}&6.9735\\
\bottomrule
\end{tabular}
}
\vspace{-0.1in}
\end{table*}

\vspace{-0.1in}
\subsection{Baselines}
To ensure a thorough and fair evaluation, we compare TimeReasoner against a diverse set of strong baselines covering three major categories:
(i) Deep learning-based models, which represent state-of-the-art architectures for time series forecasting, including DLinear~\cite{zeng2023transformers}, PatchTST~\cite{nie2022time}, iTransformer~\cite{liu2023itransformer}, and TimeXer~\cite{wang2024timexer};
(ii) LLM-based approaches, which leverage large language models for time series analysis, including GPT4TS~\cite{zhou2023one}, Time-LLM~\cite{jin2023time}, and LLMTime~\cite{gruver2023large};
(iii) Time series foundation models, which aim to provide general-purpose forecasting capability across domains, including MOIRAI~\cite{woo2024unified}, MOMENT~\cite{goswami2024moment}, and Chronos~\cite{ansari2024chronos}.

\subsection{Implementation Details}  
The TimeReasoner is model-independent; in this work, we adopt DeepSeek-R1\footnote{https://huggingface.co/deepseek-ai/DeepSeek-R1} for its open-source availability and ability to expose complete reasoning traces.
By default, all experiments are conducted under the one-shot reasoning paradigm, with inference performed via the official API\footnote{https://platform.deepseek.com} using temperature=0.6, top-p=0.7, and max-tokens=8,192. Foundation model baselines follow the same zero-shot inference protocol, while deep learning baselines are trained using the Adam optimizer~\cite{kingma2014adam} in PyTorch~\cite{paszke2019pytorch} on a single NVIDIA GeForce RTX 4090D GPU, with official implementations and recommended parameters. The lookback window is set to \(L = 36\) for NASDAQ, \(L = 300\) for VitalDB, and \(L = 96\) for other datasets; the predict window is \(H = 36\), \(H = 200\), and \(H = 96\) respectively. We adopt a channel-independent evaluation. Following standard TSF practice, we report mean squared error (MSE) to measure squared deviations and mean absolute error (MAE) to measure absolute deviations between predictions and ground truth. 

\vspace{-0.1in}
\section{Experimental Results}
In this section, we conduct a comprehensive set of experiments to empirically evaluate the performance of TimeReasoner and validate the potential of slow-thinking for time series forecasting.

\vspace{-0.2in}
\subsection{Forecasting Performance Comparison}
\textbf{To address RQ1}, we evaluate TimeReasoner’s forecasting performance across multiple benchmarks against state-of-the-art baselines.
Table~\ref{main_result} summarizes the forecasting results of TimeReasoner, demonstrating its competitive performance over other baseline models.
Results on ten datasets show that TimeReasoner achieves competitive forecasting results on most datasets, especially on ETTh1 and AQShunyi. We observe that TimeReasoner particularly excels on datasets characterized by complex temporal dynamics. On datasets like AQWan, AQShunyi and Wind, TimeReasoner consistently surpasses other methods by effectively capturing underlying dependencies. These findings underscore TimeReasoner’s outstanding capability in tackling difficult forecasting tasks and its generalization to complex time series. We also observe that the model achieves performance comparable to strong baselines on other datasets. This suggests that the proposed approach does not introduce performance degradation outside scenarios involving complex temporal dynamics.

\begin{table*}[h]
	\centering
	\caption{Ablation study on timestamps, context information and normalization strategies.}
    \vspace{-0.12in}
	\renewcommand{\arraystretch}{1}
	\normalsize
	\resizebox{\textwidth}{!}{
		\begin{tabular}{llcccccc}
			\toprule
			\multirow{2}{*}{Key Settings} & \multirow{2}{*}{Models} & \multicolumn{2}{c}{ETTh1} & \multicolumn{2}{c}{Wind} & \multicolumn{2}{c}{NASDAQ} \\
			\cmidrule(r{2pt}){3-4}
			\cmidrule(l{2pt}r{2pt}){5-6}
			\cmidrule(l{2pt}){7-8}
			&  & MSE & MAE & MSE & MAE & MSE & MAE \\
			\midrule
			\cellcolor{lightgray!30}Components & \cellcolor{lightgray!30}TimeReasoner  & \cellcolor{lightgray!30}\textbf{5.4469}       & \cellcolor{lightgray!30}\textbf{1.1984}     & \cellcolor{lightgray!30}\textbf{1556.5316}       & \cellcolor{lightgray!30}\textbf{17.5034}     & \cellcolor{lightgray!30}\textbf{0.0008}    & \cellcolor{lightgray!30}\textbf{0.0215}   \\
			\midrule
			\multirow{3}{*}{\begin{tabular}[l]{@{}l@{}}Timestamps \end{tabular}} 
			&\textit{w/o} timestamps & 25.3058      & 2.4705     & 3278.3356      &29.7851    & 0.0021    & 0.0476   \\
			&\textit{w/} forward shifting      & 7.9222       & 1.3044     &     1570.5614   & 17.8587     & 0.0009    & 0.0218   \\
			&\textit{w/} backward shifting & 6.4596      & 1.2640     & 2195.4411      & 19.3174     & 0.0010    & 0.0218   \\
			\midrule
			\multirow{1}{*}{Context Information} 
			&\textit{w/o} context information  & 7.8781       & 1.4325     & 2239.5041       & 19.4609     & 0.0009    & 0.0238   \\
			\midrule
			\multirow{2}{*}{Normalization} 
			&\textit{w/} Z-score Normalization  & 8.7691       & 1.4601     & 2076.8085       & 18.8674     & 0.0011    & 0.0236   \\
			&\textit{w/} RevIN  &  122.0912     &  11.7891    &  6641.7854      & 67.1268     &  0.0025  &   0.0671\\
			\bottomrule
		\end{tabular}
	}
	\label{ablation}
    \vspace{-0.12in}
\end{table*}

\vspace{-0.1in}
\subsection{Analysis of Key Settings in TimeReasoner}
\textbf{To address RQ2}, we conduct ablations on key components of TimeReasoner, including hybrid prompt constructions and reasoning strategies, to assess their impact on accuracy.
\vspace{-0.1in}
\subsubsection{Impact of Hybrid Prompt}
We study how prompt-design choices affect TimeReasoner’s forecasting performance by ablating three components: timestamps, contextual information, and normalization, as shown in Table~\ref{ablation}.
For timestamps, we consider two perturbations: removal, which discards all timestamp annotations, and shifting, which applies a uniform forward/backward offset while preserving relative intervals. Shifting results in only marginal degradation, suggesting robustness to changes in calendar-time anchoring, whereas removal causes a pronounced drop, highlighting the importance of temporal cues.

For contextual information, removing domain-specific descriptors degrades performance on datasets with strong seasonal or location patterns (e.g., Wind) but has little impact on highly volatile series (e.g., NASDAQ), indicating that the benefit of context is dataset-dependent. For normalization, we compare Z-score and RevIN with using raw time series and find that unnormalized inputs perform best. This suggests that standard normalization, while often helpful for deep forecasting models, can attenuate salient structures (e.g., trends and abrupt changes) that are useful for slow-thinking reasoning, and that preserving the original scale may be more compatible with our prompting setup.

\begin{figure}[t]
	\centering
	\includegraphics[width=0.48\textwidth]{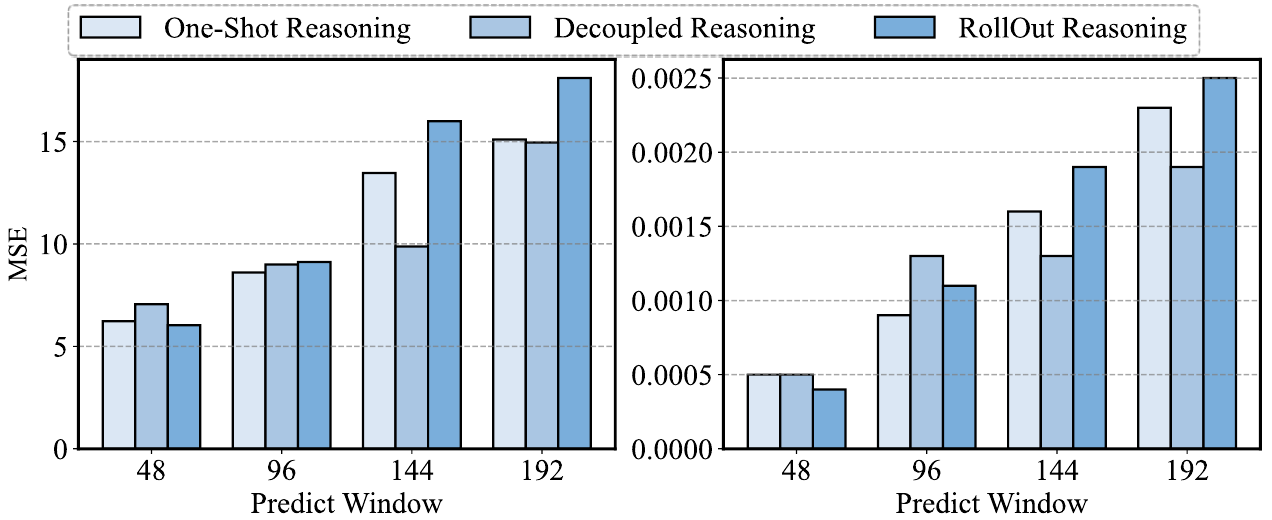}
    \vspace{-0.2in}
	\caption{Comparison of TimeReasoner’s forecasting capabilities under the three reasoning strategies with experiments conducted on ETTh2 (left) and Exchange (right) datasets.}
    \Description{Comparison of TimeReasoner’s forecasting capabilities under the three reasoning strategies with experiments conducted on ETTh2 (left) and Exchange (right) datasets.}
	\label{reasoning}
    \vspace{-0.2in}
\end{figure}

\subsubsection{Impact of Various Reasoning Strategies}
We hypothesize that Rollout Reasoning may be more effective for short-horizon forecasting, as multi-round prediction can limit error propagation when the horizon is small, while Decoupled Reasoning may be better suited for long-horizon forecasting by enabling more thorough refinement within a single inference pass. We fix the lookback window to $L=96$ and evaluate four predict windows $H \in \{48,96,144,192\}$.

Figure~\ref{reasoning} shows that One-Shot Reasoning remains competitive across all horizons. Rollout Reasoning performs best at short horizons ($H=48,96$), but its error grows more rapidly as the horizon increases ($H=144,192$), consistent with error accumulation from iterative rollouts. In contrast, Decoupled Reasoning becomes the best-performing method at longer horizons, suggesting that its within-pass refinement yields more stable predictions when long-range dependencies dominate.

\begin{table}
	\centering
	\caption{Experimental results on ETTh1 and ETTm1 datasets showing model stability under different input settings.}
    \vspace{-0.12in}
	\label{missing}
	\begin{tabular}{@{}lcccc@{}}
		\toprule
		\multirow{2}{*}{Variants} & \multicolumn{2}{c}{ETTh1} & \multicolumn{2}{c}{ETTm1} \\
		\cmidrule(lr){2-3} \cmidrule(lr){4-5}
		& MSE & MAE & MSE & MAE \\
		\midrule
		\cellcolor{lightgray!30} Full  & \cellcolor{lightgray!30}\textbf{5.4469} & \cellcolor{lightgray!30}\textbf{1.1984} & \cellcolor{lightgray!30}\textbf{14.2055} & \cellcolor{lightgray!30}\textbf{1.6984} \\
		\midrule
	 No-Imp & 9.4523 & 1.4781 & 24.4894 & 2.2308 \\
		 None-Imp & 6.6059 & 1.2498 & 14.5843 & 1.7418 \\
		Lin-Imp & \underline{5.7765} &\underline{1.2218} &\underline{14.4251} & \underline{1.7154} \\
		\bottomrule
	\end{tabular}
    \vspace{-0.2in}
\end{table}

\begin{figure*}[t]
	\centering
	\includegraphics[width=\textwidth]{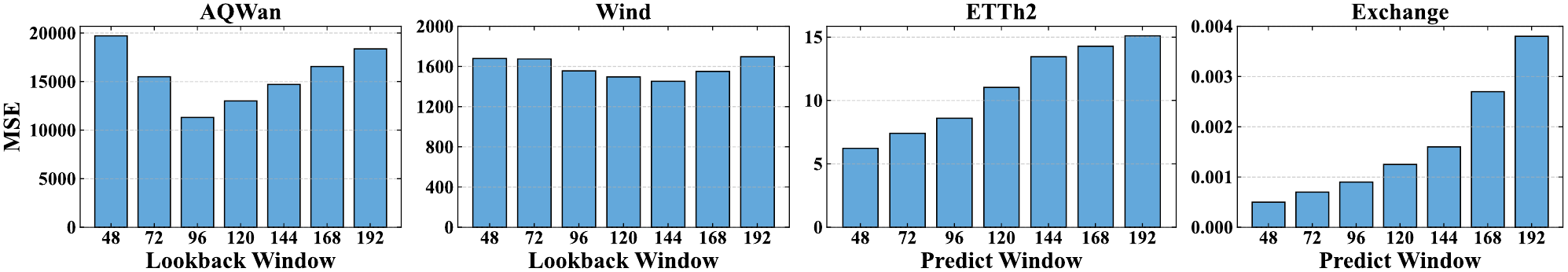}
    \vspace{-0.2in}
	\caption{Relationship between TimeReasoner's forecasting performance and lookback window and predict window.}
    \Description{Relationship between TimeReasoner's forecasting performance and lookback window and predict window.}
	\label{lookback_predict}
    \vspace{-0.2in}
\end{figure*}

\vspace{-0.1in}
\subsection{Trustworthiness Evaluation}
To address RQ3, we assess TimeReasoner’s robustness by simulating missing and noisy inputs, and further evaluate its uncertainty estimation capability.
\subsubsection{Stability Evaluation w.r.t. Missing Input}
We evaluate the stability of TimeReasoner under missing values, a common challenge in real-world time series (e.g., due to sensor failures~\cite{che2018recurrent}).
While traditional models typically rely on imputation, slow-thinking LLM-based approaches can process raw, incomplete data directly. We consider three variants: (1) No Imputation (No-Imp), where missing entries are removed; (2) None Placeholder Imputation (None-Imp), where missing values are replaced with a \texttt{None} token; and (3) Linear Interpolation Imputation (Lin-Imp)~\cite{chow1971best}.  As shown in Table~\ref{missing}, TimeReasoner performs best with fully observed data, demonstrating its ability to capture complete temporal dependencies. It remains robust under linear interpolation, while using \texttt{None} placeholders causes moderate degradation due to increased uncertainty. In contrast, dropping missing entries leads to a sharp decline, as it disrupts the temporal structure. These results suggest that preserving temporal continuity is more beneficial than deleting missing entries even when using placeholders.

\begin{figure}[t]
	\centering
	\includegraphics[width=0.48\textwidth]{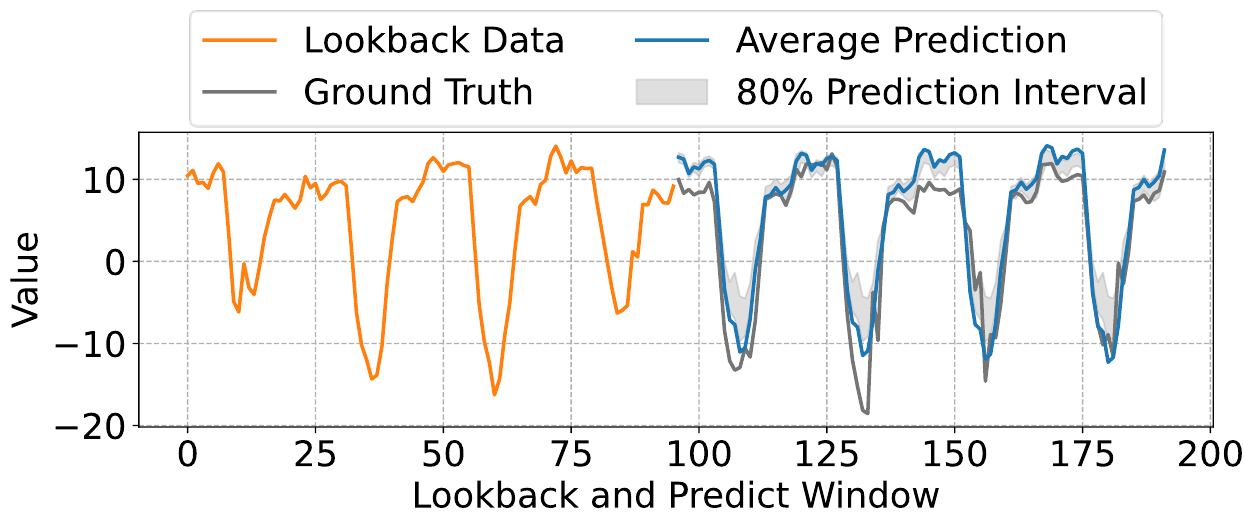}
    \vspace{-0.2in}
	\caption{The 80\% confidence interval of prediction and average prediction results during 50 independent generations given the same input on ETTh1 dataset.}
    \Description{The 80\% confidence interval of prediction and average prediction results during 50 independent generations given the same input on ETTh1 dataset.}
	\label{uncertain}
    \vspace{-0.2in}
\end{figure}

\subsubsection{Evaluation of Forecasting Uncertainty}
In this section, we aim to evaluate TimeReasoner’s capability in forecasting uncertainty quantification, with the goal of examining how well it captures the inherent reasoning uncertainty of large language models during multi-step inference and assessing both the accuracy and reliability of its probabilistic outputs.

\begin{figure}[t]
	\centering
	\includegraphics[width=0.48\textwidth]{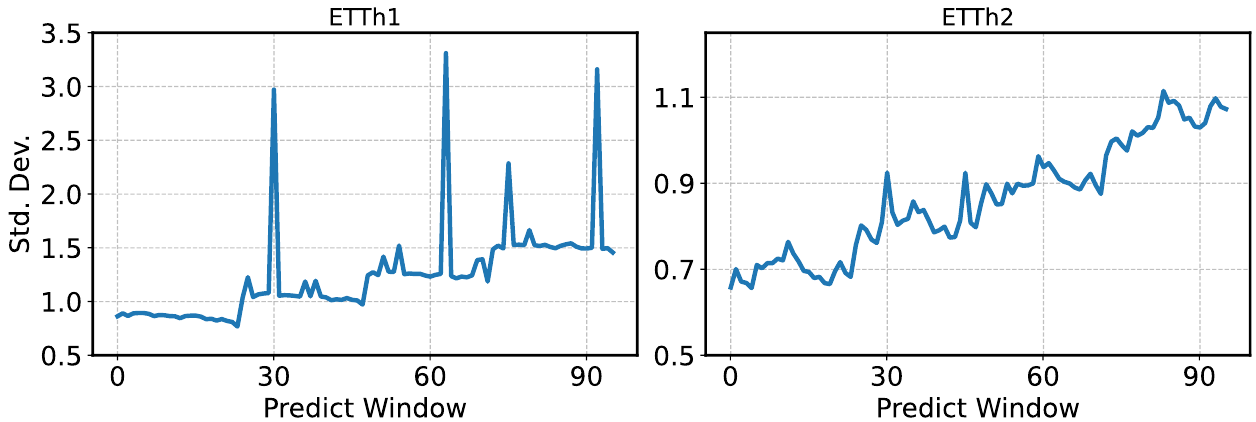}
    \vspace{-0.2in}
	\caption{Standard deviation of predictions at each forecast step across ETTh1 and ETTh2 datasets.}
    \Description{Standard deviation of predictions at each forecast step across ETTh1 and ETTh2 datasets.}
	\label{uncertain2}
	\vspace{-0.2in}
\end{figure}

As shown in Figure~\ref{uncertain}, predictions from the same LLM given identical input can vary notably, with the 80\% prediction interval spanning a wide range across the forecast horizon. This highlights the inherent uncertainty in LLM-based forecasting and underscores the need to account for it when interpreting outputs. To quantify the reasoning uncertainty in slow-thinking forecasting, we compute the standard deviation of multiple predictions at each forecast step. Figure~\ref{uncertain2} shows per-step uncertainty curves across four datasets, each line representing the deviation trajectory over the forecast window. Uncertainty generally increases with forecast length, especially on ETTh1 and ETTh2, suggesting accumulation over longer generations. Sharp spikes—most notably on ETTh1 and ETTm1 reflect moments of high uncertainty. These patterns underscore the stochastic nature of LLM-based forecasting and the need for fine-grained uncertainty estimation.

\vspace{-0.1in}
\subsection{Evaluation Under Different TSF Settings}
\textbf{To address RQ4}, we vary the look-back window and prediction length to assess TimeReasoner’s adaptability across different TSF task configurations.
\subsubsection{Performance Comparison w.r.t Varying Lookback Window.}
To examine how the lookback window influences TimeReasoner’s performance---given that longer histories often benefit traditional time series forecasting---we fix the predict window at $H=96$ and vary $L\in\{48,72,96,120,144,168,192\}$. As shown in Figure~\ref{lookback_predict}, performance follows a non-monotonic trend: increasing $L$ initially improves accuracy by providing richer patterns and context, but beyond an optimal length it degrades due to irrelevant or noisy information. This suggests that while TimeReasoner benefits from historical context, excessively long inputs can dilute salient signals, making it essential to balance context richness and relevance.

\subsubsection{Performance Comparison w.r.t Varying Predict Window}
In previous approaches, forecasting errors generally exhibit a positive correlation with the predict window length H since longer horizons introduce greater error propagation. Whether this principle still holds for TimeReasoner, which relies on step-by-step slow reasoning rather than learned representations, remains unclear. To investigate this, we fix the lookback window L = 96 and gradually increase the predict window $H\in\left\{48,72,96,120,144,168,192\right\}$. As shown in Figure \ref{lookback_predict}, MSE increases with a longer predict window. This suggests that the previous principle is applicable to TimeReasoner. This performance degradation suggests that TimeReasoner’s reasoning becomes more difficult since it extrapolates further from observed data, limiting its effectiveness for long-term forecasting.

\vspace{-0.1in}
\subsection{Hyperparameter Sensitivity Analysis}
\textbf{To address RQ5}, we analyze the effects of inference-time hyperparameters, such as temperature and base LLM selection, on forecasting performance and stability.

\subsubsection{Performance Comparison under Varying Temperatures.}
\begin{figure}[t]
	\centering
	\includegraphics[width=0.48\textwidth]{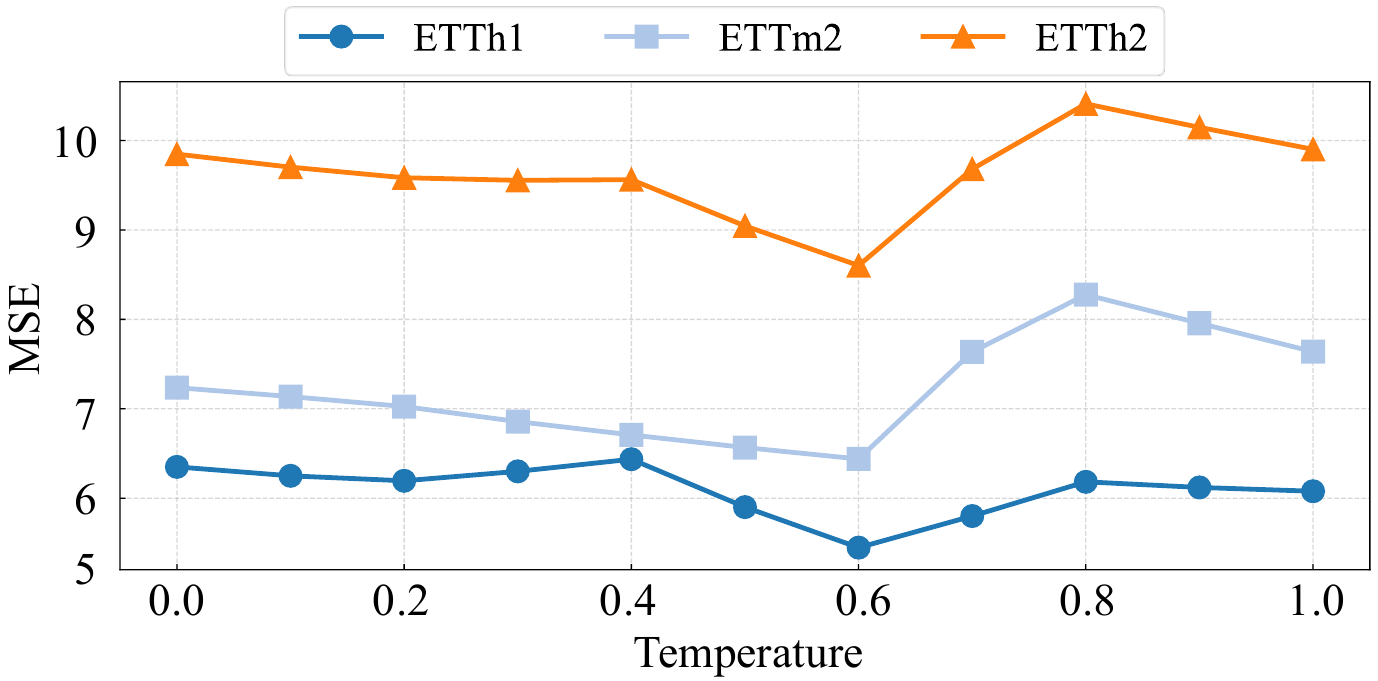}
	\vspace{-0.2in}
	\caption{Sensitivity analysis of TimeReasoner to the temperature parameter across ETTh1, ETTh2, and ETTm2 datasets.}
    \Description{Sensitivity analysis of TimeReasoner to the temperature parameter across ETTh1, ETTh2, and ETTm2 datasets.}
	\label{temperature}
    \vspace{-0.2in}
\end{figure}

Temperature controls the randomness of token sampling during LLM decoding, thereby affecting output determinism and variability. To quantify its impact on TimeReasoner, we vary the decoding temperature while keeping the prompt construction and all other settings (including the lookback window and predict window) fixed, and evaluate forecasting performance. As shown in Figure~\ref{temperature}, MSE exhibits a non-monotonic pattern: starting from $\tau=0$, performance improves and achieves the lowest MSE at $\tau=0.6$, but degrades as $\tau$ increases further. Lower temperatures yield more stable yet less flexible generations that may miss subtle temporal variations, whereas higher temperatures introduce excessive randomness, increasing variance and reducing reliability. These results suggest that an intermediate temperature provides a better trade-off between stability and controlled diversity in generation.

\begin{figure}[t]
	\centering
	\includegraphics[width=0.48\textwidth]{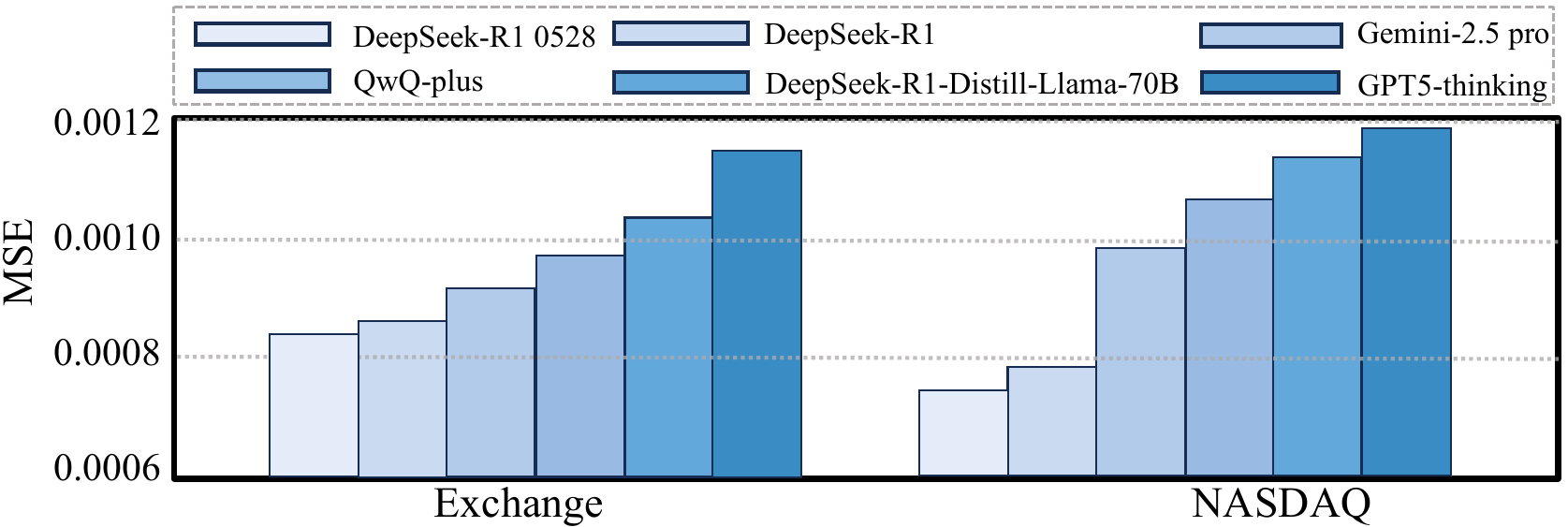}
    \vspace{-0.2in}
	\caption{Comparative performance of TimeReasoner with different base LLMs on Exchange and NASDAQ datasets.}
    \Description{Comparative performance of TimeReasoner with different base LLMs on Exchange and NASDAQ datasets.}
    \vspace{-0.2in}
	\label{llm}
\end{figure}

\subsubsection{Performance Comparison across Base LLMs}
We compare TimeReasoner across six backbone reasoning LLMs: DeepSeek-R1-0528, DeepSeek-R1, Gemini-2.5 Pro, QwQ-plus, DeepSeek-R1-Distill-Llama-70B, and GPT-5-Thinking. As shown in Figure~\ref{llm}, DeepSeek-R1-0528 delivers the best forecasting performance on both Exchange and NASDAQ, followed closely by DeepSeek-R1, while the distilled variant performs worst, highlighting the performance gap introduced by distillation. The improvement from DeepSeek-R1 to DeepSeek-R1-0528 further indicates that TimeReasoner benefits from stronger backbone reasoning capabilities.

Interestingly, GPT-5-Thinking underperforms in this setting. Qualitative inspection suggests that it sometimes relies on simple code-style extrapolation heuristics (e.g., linear or mean extrapolation) instead of closely tracking distributional cues in the lookback window, which can reduce responsiveness to regime shifts and degrade accuracy. Overall, these results suggest that the \emph{style} of reasoning—data-attentive inference versus heuristic shortcutting—can be as important as model scale, and that observation-first prompting and data-focused exemplars may be a promising direction for improving robustness.

\vspace{-0.1in}
\subsection{Visualization Analysis}
\textbf{To address RQ6}, we analyze intermediate reasoning traces and a representative forecasting case to assess whether the output is aligned with the input data.
\subsubsection{Reasoning Trace Analysis}  
Building on TimeReasoner’s zero-shot forecasting performance, we examine its model-generated CoT-style reasoning traces. We observe a common three-stage structure: (1) Data inspection and pattern identification, where the model summarizes salient temporal characteristics (e.g., periodicity/seasonality, trend, and level shifts); (2) Strategy selection and justification, where it evaluates candidate forecasting heuristics (from seasonal baselines to trend extrapolation) and discusses their trade-offs; and (3) Self-reflection and contextual checks, where it critiques the chosen strategy, noting strengths (e.g., capturing seasonality) and limitations (e.g., lacking explicit statistical mechanisms). Overall, these traces suggest a diagnostic--select--verify workflow prior to producing the final forecast.

\begin{figure}[t]
    \centering
    \includegraphics[width=0.48\textwidth]{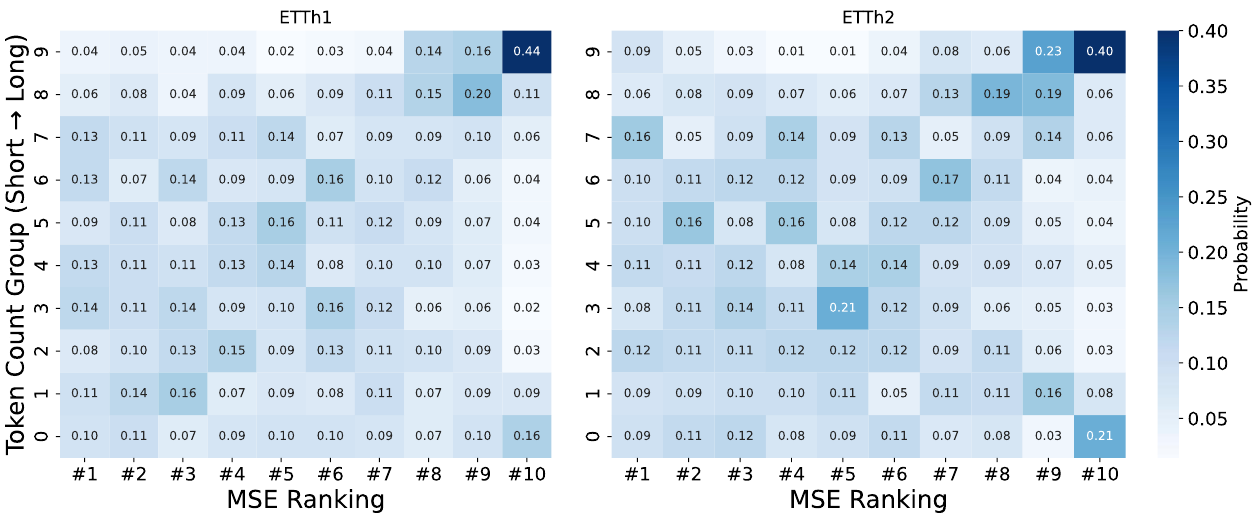}
    \vspace{-0.2in}
    \caption{Heatmap of MSE rankings across CoT length groups on ETTh1 and ETTh2. The x-axis shows MSE ranks (\#1–\#10), and the y-axis shows CoT token deciles from shortest (bottom) to longest (top).}
    \Description{Heatmap of MSE rankings across CoT length groups on ETTh1 and ETTh2. The x-axis shows MSE ranks (\#1–\#10), and the y-axis shows CoT token deciles from shortest (bottom) to longest (top).}
    \label{relitu}
    \vspace{-0.2in}
\end{figure}

We further investigate the influence of the generated Chain-of-Thought (CoT) length on forecasting accuracy. As shown in Figure~\ref{relitu}, all predictions are grouped into deciles according to their CoT token counts, with shorter CoTs placed in the lower deciles and longer ones in the upper deciles. A distinct pattern emerges: longer CoT generations are disproportionately concentrated in the worst MSE rankings, whereas shorter CoTs exhibit a more diverse distribution that occasionally includes top-ranked predictions. This observation implies that while moderate reasoning depth can aid pattern recognition and improve temporal alignment, excessively long reasoning chains may introduce irrelevant details, overfit transient patterns, or drift away from salient signals in the input. Such overextension not only inflates computational cost but can also amplify stochastic variability in token-by-token generation, ultimately degrading predictive performance. These findings highlight the importance of calibrating reasoning length, balancing depth with focus to achieve reliable and efficient LLM-based forecasting.

\subsubsection{Forecasting Case Analysis}
\begin{figure*}[t]
	\centering
	\includegraphics[width=\textwidth]{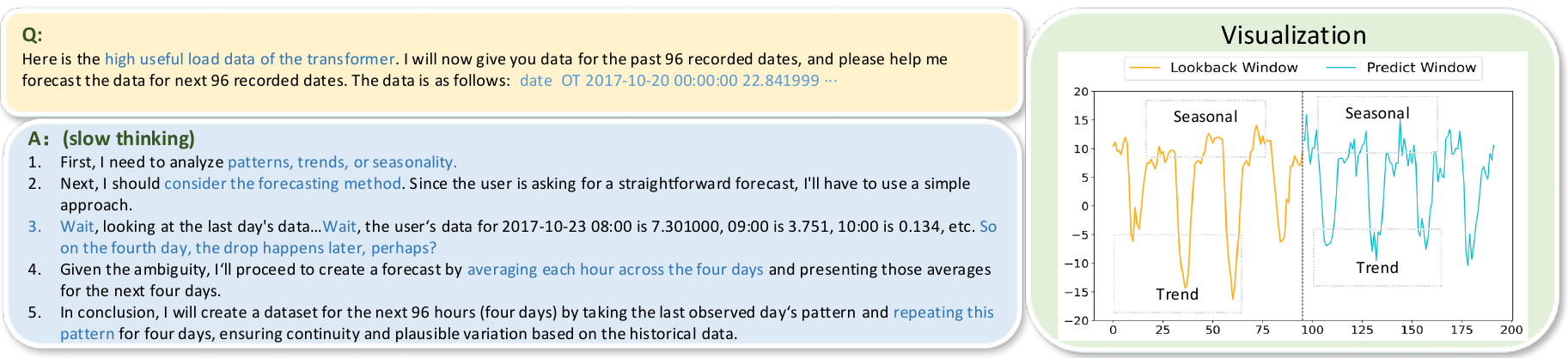}
    \vspace{-0.2in}
	\caption{A case demonstrating how TimeReasoner performs slow-thinking capability for 96-96 time series reasoning.}
    \Description{A case demonstrating how TimeReasoner performs slow-thinking capability for 96-96 time series reasoning.}
	\label{case}
    \vspace{-0.1in}
\end{figure*}

Figure~\ref{case} presents a representative example of TimeReasoner’s inference-time reasoning, illustrating its deliberative inference behavior for time-series forecasting with lookback window $L=96$ and predict window $H=96$. The model first summarizes salient temporal patterns (e.g., trend and seasonality) and then evaluates candidate forecasting strategies. It subsequently inspects the most recent observations, noting potential anomalies and regime shifts. Based on these cues, TimeReasoner forms a forecasting plan by referencing comparable historical periods and enforcing continuity across the predicted sequence. The visualization on the right shows that the predicted window preserves both seasonal and trend components observed in the lookback window, suggesting that the model can produce interpretable, step-by-step forecasts.

\begin{figure*}[t]
    \centering
    \includegraphics[width=\textwidth]{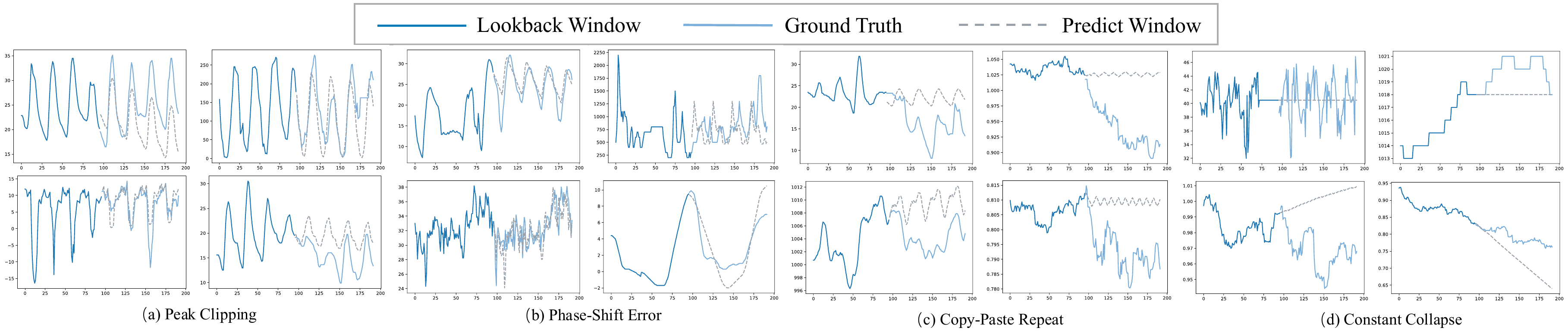}
    \vspace{-0.2in}
    \caption{Four representative failure modes of TimeReasoner, including (a) Peak Clipping, (b) Phase-Shift Error, (c) Copy-Paste Repeat, and (d) Constant Collapse.}
    \Description{Four representative failure modes of TimeReasoner, including (a) Peak Clipping, (b) Phase-Shift Error, (c) Copy-Paste Repeat, and (d) Constant Collapse.}
    \label{abnormal_case}
    \vspace{-0.12in}
\end{figure*}

\subsubsection{Failure Case Analysis}
We observe that not all forecasts generated by TimeReasoner are successful, with some exhibiting systematic and recurring errors. To better understand these limitations, we conduct a detailed manual inspection of poorly performing cases and categorize them into four representative failure patterns, as shown in Figure~\ref{abnormal_case}. (a) Peak clipping refers to systematic underestimation or overestimation at peaks and troughs, which flattens critical turning points and leads to the loss of important extreme-value information. (b) Phase-shift error describes cases where predictions capture the correct overall shape of the ground truth but are misaligned in time, causing decisions to be triggered too early or too late. (c) Copy-paste repeat reflects the tendency to directly replicate segments from the lookback window without adapting to new conditions, resulting in stale and inaccurate forecasts. (d) Constant collapse denotes degeneration into nearly constant outputs, disregarding inherent variability in the data.

These patterns reveal distinct weaknesses in amplitude capture, temporal alignment, adaptability, and output diversity. Such errors can severely undermine forecast reliability—misaligned peaks may cause decision failures in time-critical applications, excessive repetition can overlook regime changes or unexpected events, and constant predictions fail to reflect dynamic trends, leading to missed opportunities or inadequate responses. Beyond lowering pointwise accuracy, these issues also reduce the explainability and trustworthiness of the model’s reasoning process. Addressing them will require enhancing sensitivity to extremes, improving phase anchoring to the correct temporal context, and discouraging trivial or degenerate outputs, thereby ensuring that the model remains both accurate and responsive across diverse forecasting scenarios.

\vspace{-0.1in}
\section{Conclusion and Discussions}
This work explored whether large language models (LLMs) with slow-thinking capabilities can reason over temporal patterns for time series forecasting. We reformulated forecasting as a conditional reasoning task, encoding historical time series and contextual features into structured prompts to guide trajectory generation.
To realize this paradigm, we proposed TimeReasoner, a prompting-based framework that integrates hybrid inputs, including time series, timestamps, and semantic descriptors, further supporting multiple inference-time reasoning strategies (one-shot, decoupled, and roll-out).
Experiments across diverse TSF benchmarks demonstrate that slow-thinking LLMs achieve strong zero-shot forecasting performance, while their intermediate reasoning traces reveal coherent patterns of temporal dynamics. This study represents an initial step toward reasoning-centric forecasting systems. 

\noindent\textbf{Discussions.} Our results demonstrate the potential of slow-thinking approaches for time series forecasting, showing that even a relatively simple framework like TimeReasoner can benefit from the evolving ecosystem of reasoning-oriented LLMs, such as DeepSeek-R1, and is expected to gain further improvements as more advanced models become available. However, challenges remain: the reasoning trajectories generated by LLMs still exhibit uncertainty, calling for future work on principled uncertainty quantification; and the existence of erroneous predictions in some cases highlights the need for more robust reasoning and error-correction mechanisms. Addressing these limitations will be essential for realizing reliable and generalizable reasoning-based forecasting systems.

\section{Acknowledgments}
This research was supported by grants from the National Natural Science Foundation of China (No. 62502486, 62337001), the grants of Provincial Natural Science Foundation of Anhui Province (No. 2408085QF193), USTC Research Funds of the Double First-Class Initiative (No. YD2150002501), the Fundamental Research Funds for the Central Universities of China (No. WK2150110032).

\newpage

\section*{Ethical Considerations}
All experiments in this study are conducted on publicly available, fully anonymized benchmark datasets, which contain no personal or sensitive information. The research does not involve human subjects, protected attributes, or any data that could lead to privacy violations. No foreseeable negative societal impacts are identified, such as unfair treatment, malicious surveillance, or deliberate misuse by adversarial actors. The proposed TimeReasoner framework is designed for general-purpose time series forecasting research and does not target sensitive or safety-critical application domains. The work follows established ethical guidelines for data usage, reproducibility, and responsible scientific research, and all experimental settings are documented to ensure transparency and replicability.

\bibliographystyle{ACM-Reference-Format}
\balance
\bibliography{sample-base}
\end{document}